\documentclass[letterpaper, 10 pt, journal, twoside]{IEEEtran}
\IEEEoverridecommandlockouts
\usepackage{cite}
\usepackage{amsmath,amssymb,amsfonts}
\usepackage{algorithmic}
\usepackage{graphicx}
\usepackage{textcomp}
\usepackage{xcolor}
\usepackage{pgfplots} 
\usepgfplotslibrary{polar}
\pgfplotsset{compat=1.15}
\usepackage{siunitx}
\usepackage{multirow}
\usepackage{hyperref}

\def\BibTeX{{\rm B\kern-.05em{\sc i\kern-.025em b}\kern-.08em
    T\kern-.1667em\lower.7ex\hbox{E}\kern-.125emX}}
\begin{document}

\definecolor{red}{rgb}{1.00,0.00,0.00}
\definecolor{lightred}{rgb}{1.00,0.3,0.3}
\definecolor{blue}{rgb}{0.00,0.00,1.00}
\definecolor{green}{rgb}{0.1,0.50,0.1}
\definecolor{yellow}{rgb}{0.5,0.5,0.0}
\definecolor{white}{rgb}{1,1,1}
\definecolor{gray}{rgb}{0.6,0.6,0.6}
\newcommand{\cred}[1] {\textcolor{red}{\textbf{#1}}}
\newcommand{\cblue}[1] {\textcolor{blue}{\textbf{#1}}}
\newcommand{\cgreen}[1] {\textcolor{green}{#1}}
\newcommand{\cyellow}[1] {\textcolor{yellow}{#1}}
\newcommand{\cwhite}[1] {\textcolor{white}{#1}}
\newcommand{\cgray}[1] {\textcolor{gray}{#1}}

\title{\LARGE \bf
Learning Hybrid Locomotion Skills -- Learn to Exploit Residual Dynamics and Modulate Model-based Gait Control}



\author{Mohammadreza Kasaei$^{1}$, Miguel Abreu$^{2}$, Nuno Lau$^{3}$, Artur Pereira$^{3}$, Lu\'is Paulo Reis$^{2}$ and Zhibin Li$^{1}$%

\thanks{$^{1}$Mohammadreza Kasaei and Zhibin Li are with School of Informatics, University of Edinburgh, UK
        {\tt\footnotesize \{m.kasaei, zhibin.li\}@ed.ac.uk}}%
\thanks{$^{2} $Miguel Abreu and  Lu\'is Paulo Reis are with LIACC/FEUP, Artificial Intelligence and Computer Science Lab, University of Porto, Portugal
        {\tt\footnotesize \{m.abreu,lpreis\}@fe.up.pt}}%
\thanks{$^{3}$Nuno Lau and Artur Pereira are with IEETA/DETI, University of Aveiro, Portugal
        {\tt\footnotesize \{nunolau, artur\}@ua.pt}}%
}
	
\maketitle

\begin{abstract} 
This work aims to combine machine learning and control approaches for legged robots, and developed a hybrid framework to achieve new capabilities of balancing against external perturbations.
The framework embeds a kernel which is a fully parametric closed-loop gait generator based on analytical control. On top of that, a neural network with symmetric partial data augmentation learns to automatically adjust the parameters for the gait kernel and to generate compensatory actions for all joints as the residual dynamics, thus significantly augmenting the stability under unexpected perturbations. The performance of the proposed framework was evaluated across a set of challenging simulated scenarios. The results showed considerable improvements compared to the baseline in recovering from large external forces. Moreover, the produced behaviours are more natural, human-like and robust against noisy sensing. 
\end{abstract}


\vspace{-3mm}

%
\IEEEpeerreviewmaketitle

\section{Introduction}\label{sec:Intro}
Legged robots are extremely versatile and can be used in wide ranges of applications. Nevertheless, robust locomotion is a complex topic which still needs investigation. {Stability and safety} are essential requirements for a robot to act in a real environment. The question is: despite the legged robots' versatility, why are they not as capable as us yet?

To achieve the versatility as intended, we investigated the fundamental aspect of learning balance recovery strategies. Humans combine a set of strategies (e.g. moving arms, ankles, hips, taking a step, etc.) to regain the balance after facing an external disturbance. They rely on past experiences to improve their methods. 
Moreover, we investigated existing biped robot locomotion frameworks. Despite their stability have been improved significantly but they are not stable and safe enough to be utilised in our daily-life environments. 
Several approaches for stabilising a biped robot have been proposed that can be categorised into three major categories. In the remainder of this section, these categories will be introduced and some recent works in each category will be briefly reviewed.
\begin{figure}[!t]
	\centering
	\includegraphics[width = 0.97\columnwidth, trim= 0.0cm 0.0cm 0.1cm 0cm,clip]{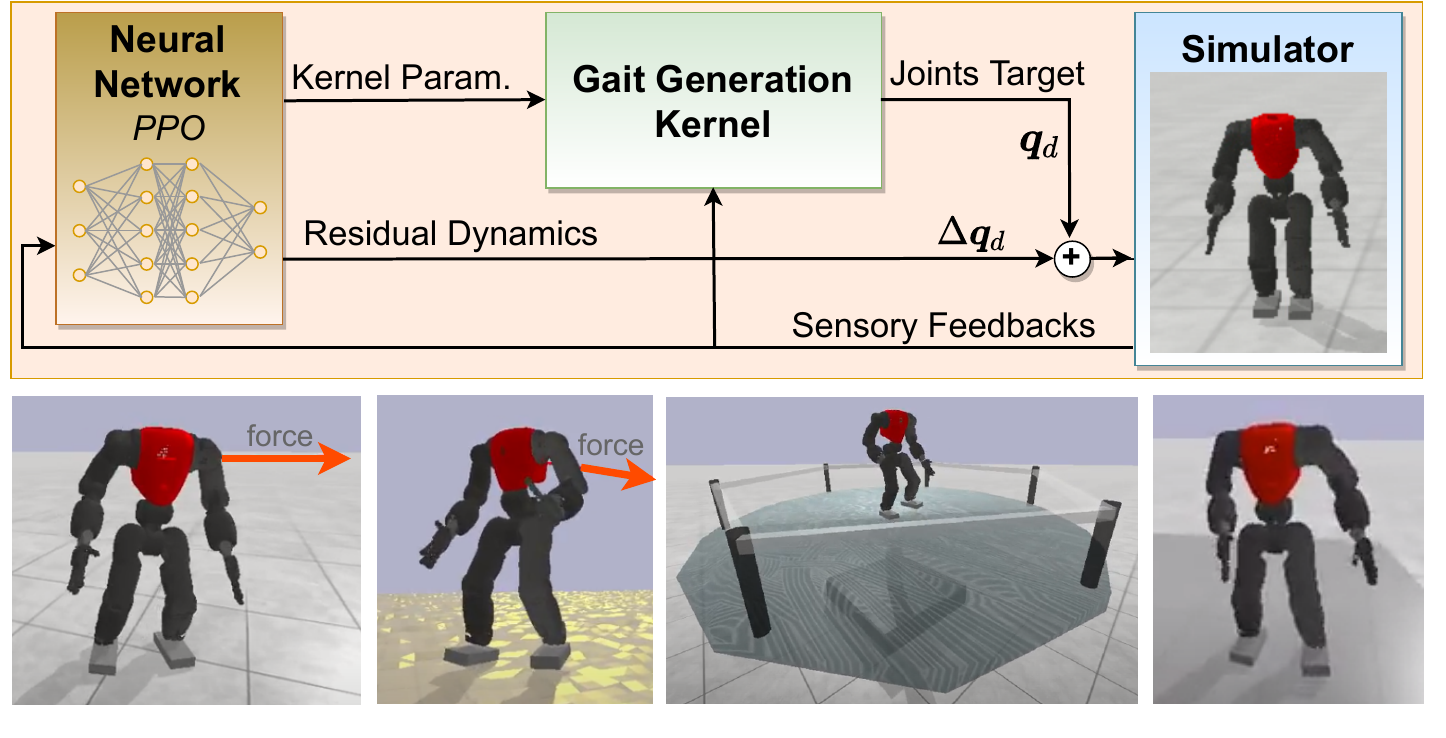}
	\vspace{-3mm}
	\caption{Overview of the proposed framework along with a set of snapshots of tests on different terrains: the gait generation kernel produces closed-loop locomotion, the neural network regulates the kernel's parameters and generates compensatory actions.}
	\vspace{-4mm}
	\label{fig:Cover}
\end{figure}

\subsection{Analytical Approaches}
The basic idea behind the approaches in this category is using a dynamics model of the robot and designing a set of controllers (e.g., force controller~\cite{mason2016balancing}, hybrid position/force~\cite{faraji2019bipedal}, admittance controller~\cite{caron2020icra}) based on some criteria to minimise the tracking error. The most widely used model in literature is the Linear Inverted Pendulum~(LIP) which abstracts the overall dynamics of a robot as a single mass. It restricts the vertical movement of the mass to provide a linear model which yields a fast solution for real-time implementations.  This model has been investigated and extended for decades to design and analyse legged robot locomotion~\cite{takenaka2009real,englsberger2015three}.

Takaneka et al.~\cite{takenaka2009real} proposed the Divergent Component of Motion~(DCM) concept that splits the LIP's dynamics into stable and unstable parts, such that controlling the unstable part is enough for keeping the stability. In~\cite{englsberger2015three}, DCM has been extended to 3D and, several control approaches including classical feedback controllers~\cite{morisawa2014biped}, Linear Quadratic Regulator~(LQR)-based methods~\cite{faraji2019bipedal,8967778} and the Model Predictive Control~(MPC)~\cite{ brasseur2015robust} have been proposed to formulate biped locomotion frameworks. All of them are trying to compensate the tracking error by using a combination of three strategies which are: manipulating the Ground Reaction Force~(GRF) and modifying the position and time of the next step.

Recently, researchers are investigating the release of LIP assumptions (e.g. COM vertical motion and angular momentum) which causes dealing with nonlinearities~\cite{caron2020icra, kajita2018biped, seyde2018inclusion}. 

\vspace{-2mm}
\subsection{Machine Learning Approaches}

The approaches in this category are designed to learn a feasible policy through interaction with the environment. Nowadays, Deep Reinforcement Learning~(DRL) has shown its capability by solving complex locomotion and manipulation tasks, which are generally composed of high-dimensional continuous observation and action spaces~\cite{gu2017deep,abreu2019learning}.

Data augmentation in DRL is widely used to improve the optimization performance but, in this work, we restrict the scope to symmetry oriented solutions. The process of generating symmetric data from actual samples is used to improve different robotic tasks~\cite{lin2020robotics}, including the walking gait of various humanoid models~\cite{abdolhosseini2019learning} and quadruped robots~\cite{mishra2019augmenting} (with more than one plane of symmetry). Learning from scratch with DRL can achieve very efficient behaviours, even in asymmetrical configurations~\cite{Abreu2019runfaster}. However, if not regulated through model restrictions (e.g. symmetry, pattern generators), it can be challenging to produce human-like behaviours in a reasonable amount of time.

\subsection{Hybrid Approaches: combing analytical and learning}
The approaches in this category are focused on combining the potential of both aforementioned categories. To do so, learning algorithms can be combined with model-based gait pattern generators to predict the parameters and to learn residual dynamics (compensatory action), which can lead to impressively accurate behaviours~\cite{koryakovskiy2018model,ahn2020data}. 

These frameworks are generally composed of a set of layers that are connected together in hierarchical structures. Yang et al.~\cite{yang2018learning} designed a hierarchical framework to ensure the stability of a humanoid robot by learning motor skills. Their framework is composed of two independent layers, the high-level layer generates a set of joint angles and the low-level layer translates those angles to joint torques using a set of PD controllers. Their reward function was composed of six distinct terms that were mostly related to the traditional push recovery strategies, and it was obtained by adding all terms together with different weights.

\subsection{Overview of the Proposed Framework and Contributions}
This work focuses on bipedal locomotion which is the most challenging in legged robots. Particularly, we aim to answer an interesting question of whether or not a learning algorithm can learn to control and modulate a model-based control policy such as a gait pattern generator. 

Our contributions are the following: \textbf{(i)} we developed a locomotion framework for humanoid robots that integrates both analytical control and machine learning. An overview of this system is depicted in Fig.~\ref{fig:Cover}. Specifically, we use an abstract dynamics model to analytically formulate a closed-loop biped locomotion and recovery strategies as a kernel, and combine it with a symmetry-enhanced optimisation framework using Proximal Policy Optimisation~(PPO)~\cite{schulman2017ppo} to learn residual dynamics. The learned policy adjusts a set of parameters of the pattern generator and generates compensatory actions as the residual dynamics to regain stability; \textbf{(ii)} we proposed a learning method where the data is only partially augmented, leveraging the symmetry to improve learning time and human-likeness without restricting asymmetric movements, thus widening the range of possible behaviours.

The remainder of this paper is structured as follows: In Section~\ref{sec:pattern}, the architecture of our fully parametric kennel pattern generator will be presented and each module will be explained. Afterwards, in Section~\ref{sec:learningResidual}, our learning framework will be introduced and we will explain how we augmented this framework with the  kernel  pattern  generator to  regulate  kernel  parameters and to learn model-free skills~(generating compensatory joint  positions). In Section~\ref{sec:simScenarios}, a set of simulation scenarios will be designed to validate the performance of the proposed framework. Afterwards, in Section~\ref{sec:simulations}, a bunch of simulations will be conducted to provide assessments and analysis regarding overall performance, optimized policy  behaviour, symmetry, and robustness. Finally, conclusions and future research are presented in Section~\ref{sec:conclusion}.

\section{Gait Generation Kernel}  \label{sec:pattern}
\begin{figure}[!t]
	\centering
	\includegraphics[width = 0.95\columnwidth]{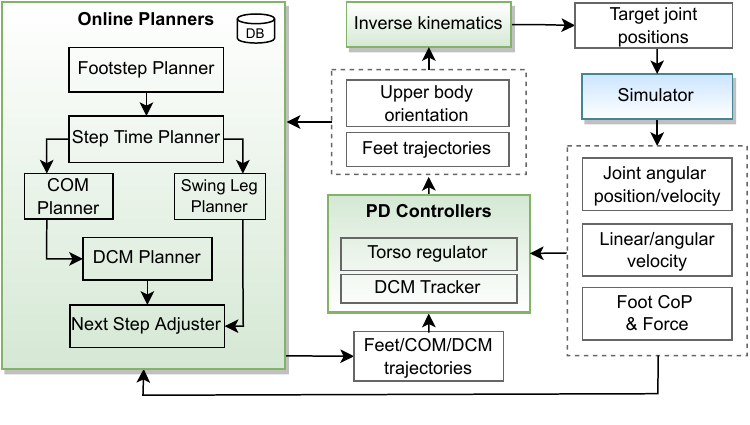}
	\vspace{-3mm}
	\caption{Overview of the proposed kernel gait generator. The online planners module generates a set of reference trajectories according to the input command and the states of the system. The PD controllers module is responsible for tracking the generated trajectories. The coloured boxes represent the main modules and the white boxes are the exchanged information among them.}
	\vspace{-5mm}
	\label{fig:Controller}
\end{figure}

A fully parametric closed-loop gait generator serves as a kernel of the walking pattern (Fig.~\ref{fig:Controller}). The gait generator is composed of two main modules: \texttt{Online Planners} and \texttt{PD Controllers}. \texttt{Online Planners} is responsible for generating the reference trajectories according to the stride's parameters provided by the user, the robot's state and the controllers' output. \texttt{PD Controllers} regulates the upper body orientation and tracks the planned trajectories to generate closed-loop locomotion. The corresponding target joint positions are generated using \texttt{Inverse Kinematics}, taking into account the kinematic feasibility. The target joint positions are fed to the \texttt{Simulator} for simulating the interaction of the robot with the environment and producing sensory data, as well as the global position and orientation of the robot. 

\subsection{Online Planners}
The \texttt{Online Planners} is based on Kasaei et al.~\cite{8967778} that will be described briefly for the sake of completeness. As shown in Fig~\ref{fig:Controller}, \texttt{Online Planners} is composed of a set of sub-planners which are solved separately and connected together hierarchically to reduce the complexity of the planning process. The planning process is started by generating a set of footsteps ($\boldsymbol{f}_i = [f_{i_x} \quad f_{i_y} ]^\top \quad i \in \mathbb{N} $) according to the input stride's parameters and the current feet configuration. Then, the step time planner assigns a set of timestamps to the generated footstep according to the stride duration. Afterwards, to have a smooth trajectory during lifting and landing of the swing foot, a cubic spline is used to generate the swing leg trajectory based on the generated footsteps and a predefined swing height. 

Accordingly, the COM planner generates the COM trajectory by solving LIP equation as a boundary value problem based on the generated footsteps. Then, the DCM trajectory can be obtained by substituting the generated COM and its time derivative into DCM equation ($\boldsymbol{\zeta} = \boldsymbol{c}+\frac{\boldsymbol{\dot{c
}}}{\omega}$, where $\boldsymbol{\zeta}$ is DCM; $\boldsymbol{c}$ and $\boldsymbol{\dot{c}}$ represent the COM and its time derivative, respectively). 
This trajectory will be fed into \texttt{PD Controllers} to generate closed-loop locomotion.
More detail can be found in our previous work~\cite{8967778}.


In some situations, such as when the robot is being pushed severely, the DCM tracker cannot track the reference because of the controllers' output saturation. In such conditions, humans adjust the next step time and location, in addition to the COM's height. Due to the observability of DCM at each control cycle, the position of the next step can be determined by solving DCM equation as an initial value problem:
\begin{equation}
\boldsymbol{f}_{i+1} = \boldsymbol{f}_{i} + (\boldsymbol{\zeta}_t - \boldsymbol{f}_{i})e^{\omega(T-t)} \;,
\label{eq:dcm_at_end}
\end{equation}
\noindent
where $\boldsymbol{f}_{i}, \boldsymbol{f}_{i+1}$ are the current and next support foot positions and $t, T$ denote the time and stride duration, respectively. 

It should be noted that adjusting the next stride time as well as the height of the COM is not straightforward due to nonlinearities. Finding optimal or near optimal values for these parameters using DRL is a desirable solution, not only due to its convergence properties, but also because it allows us to find a more complete overall strategy by combining the stride time and COM height with residual adjustments.

\subsection{Regulating the Upper Body Orientation}
The upper body of a humanoid is generally composed of several joints. While the robot is walking, their motions and vibrations generate angular momentum around the COM. To cancel the effects of this momentum, we designed a PD controller ($\boldsymbol{u}_\Phi$) based on the inertial sensor values that are mounted on the robot's torso:
\begin{equation}
\boldsymbol{u}_\Phi = -\boldsymbol{K}_{\Phi} (\boldsymbol{\Phi}_d - \boldsymbol{\Phi})  \;, 
\label{eq:PDTorso}    
\end{equation}
\noindent
where \mbox{$\boldsymbol{\Phi} = [\Phi_{roll}\quad \dot{\Phi}_{roll}\quad \Phi_{pitch}\quad \dot{\Phi}_{pitch}]^\top$} represents the state of the torso and $\boldsymbol{\Phi}_d$ denotes the desired state of the torso and $\boldsymbol{K}_\Phi$ is the controller gains.

\subsection{DCM Tracker}

According to the LIP and DCM, the overall dynamics of a humanoid robot can be can be represented by a linear state space system as follows:
\begin{equation}
\frac{d}{dt} \begin{bmatrix} \boldsymbol{c} \\ \boldsymbol{\zeta} \end{bmatrix}
= 
\begin{bmatrix} 
-\omega\boldsymbol{I}_2 & \omega\boldsymbol{I}_2  \\ 
0\boldsymbol{I}_2  & \omega\boldsymbol{I}_2

\end{bmatrix}	
\begin{bmatrix} \boldsymbol{c} \\ \boldsymbol{\zeta} \end{bmatrix}
+
\begin{bmatrix} 
\boldsymbol{0}_{2\times1}\\
-\boldsymbol{\Omega}
\end{bmatrix} \boldsymbol{p} \;,
\label{eq:statespace_zeta}
\end{equation}
\noindent
where $\boldsymbol{I}_2$ is an identity matrix of size 2, \mbox{$\boldsymbol{c}= [c_x\quad c_y]^\top$} denotes the position of the COM,  \mbox{$\boldsymbol{\zeta} = [\zeta_x \quad \zeta_y]^\top$} is the DCM, \mbox{$\boldsymbol{p} = [p_x\quad p_y]^\top $} represents the position of the ZMP and \mbox{$\omega = \sqrt{\frac{g}{c_z}}$} is the natural frequency of the pendulum, where $g$ is the gravity constant and $c_z$ represents the height of the COM and \mbox{$\boldsymbol{\Omega} = [\omega\quad \omega]^\top$}.

This system shows that the COM is always converging to the DCM, and controlling the DCM is enough to develop stable locomotion. Thus, the DCM tracker can be formulated as:
\begin{equation}
\boldsymbol{u}_\zeta =- \boldsymbol{K}_{\zeta}\boldsymbol{e}_\zeta \;,
\label{eq:dcm_tracker}
\end{equation}
\noindent
where $\boldsymbol{K}_\zeta$ represents the controller gains, \mbox{$\boldsymbol{e}_\zeta=[\boldsymbol{\zeta}_d-\boldsymbol{\zeta} ,\quad\boldsymbol{\dot{\zeta}_d}-\boldsymbol{\dot{\zeta}}]^\top$, $\boldsymbol{\zeta}_d, \dot{\boldsymbol{\zeta}_d}$} are the desired DCM and its time derivative, which are generated by the DCM planner (see Fig.~\ref{fig:Controller}).

\section{Learning Residual Dynamics}\label{sec:learningResidual}
Although the gait generator produces stable locomotion, it does not generalise well to unforeseen circumstances. This section presents our developed learning framework that can learn \textit{residual dynamics} on top of the kernel pattern generator. The objective is to regulate control parameters such as the COM height and stride time, and also learn model-free skills to generate compensatory joint actions.
\vspace{-2mm}
\subsection{Formal structure}

The PPO algorithm was chosen as the base RL algorithm due to its computational efficiency and good performance in high-dimensional environments. The learning framework extends this algorithm with symmetric data augmentation based on static domain knowledge. Like most humanoid models, the COMAN robot has reflection symmetry in the sagittal plane, which can be leveraged to reduce the learning time and guide the optimisation algorithm in creating a human-like behaviour. 

This learning problem can be formally described as a Markov Decision Process (MDP) -- a tuple $\left\langle S,A,\Psi,p,r\right\rangle$, where $S$ is the set of states, $A$ is the set of actions, $\Psi\subseteq S\times A$ is the set of admissible state-action pairs, $p(s,a,s'):\Psi \times S \rightarrow[0,1]$ is the transition function, and $r(s,a):\Psi \rightarrow {\rm I\!R}$ is the reward function. In order to reduce the mathematical model by exploiting its redundancy and symmetry, Ravindran and Barto~\cite{ravindran2001} proposed the MDP homomorphism formalism, which describes a transformation that simplifies equivalent states and actions. Let $h$ be an MDP homomorphism from $M=\left\langle S,A,\Psi,p,r\right\rangle$ to $M'=\left\langle S',A',\Psi',p',r'\right\rangle$, and $A_s$ be the set of admissible actions in state $s$. The concept of MDP symmetries is a special case of this framework where $f:S\rightarrow S'$ and $g_s:A_s\rightarrow A'_{f(s)}$ are bijective functions. An MDP isomorphism from and to the same MDP can be considered an automorphism that satisfies:
\begin{align}
p(f(s),g_s(a),f(s')) & = p(s,a,s'), \quad \forall s, s' \in S,a\in A_s, \\
\mathrm{and} \quad  r(f(s),g_s(a)) & =r(s,a), \quad \forall s \in S, a \in A_s. \label{eq:homRew2}
\end{align}

\vspace{-4mm}
\subsection{Data augmentation}

In this work, the formulated problem is optimised using PPO~\cite{schulman2017ppo}, an actor-critic algorithm that uses a clipping function to constrain the policy update directly inside the objective function, thus preventing it from being too greedy. After performing a grid search, the batch size was set to 8192 samples and the learning rate to $3\mathrm{e}{-4}$ (using a linear scheduler). For each episode, an MDP trajectory $j$ is characterised by a sequence of states, actions and rewards such that $j=\{S_0,A_0,R_0,S_1,A_1,R_1,...\}$. Each trajectory is used to produce a set of samples $k=\{\{S_0,A_0,Ad_0,V_0\},\{S_1,A_1,Ad_1,V_1\},...\}$,  where $V_i$ is obtained from the $\lambda$-return as defined by Sutton and Barto~\cite{sutton2018reinforcement}, and serves as value target for the update function; and $Ad_i$ is the generalised advantage estimate~\cite{schulman2018gae}. 

Our proposal is to partially augment data by copying and transforming a fraction of the acquired samples. Different augmentation ratios are tested in Section \ref{sec:simulations}. As an example, consider the addition of symmetrical samples with a ratio of 50\%. Following~\eqref{eq:homRew2}, each batch of samples is artificially built as $\{W_1,W_2,u(W_2),W_3,W_4,u(W_4),...\}$ where $u(W_i) = \{f(S_i),g_s(A_i),Ad_i,V_i\}$. The observations' normalisation is continuously updated by calculating the mean and standard deviation of each observation. However, both of these metrics are shared among the two symmetric groups to ensure that no asymmetrical bias is introduced.

\begin{figure}[!t]
    \vspace{0.cm}
    \centering
    \includegraphics[width = 0.95\columnwidth]{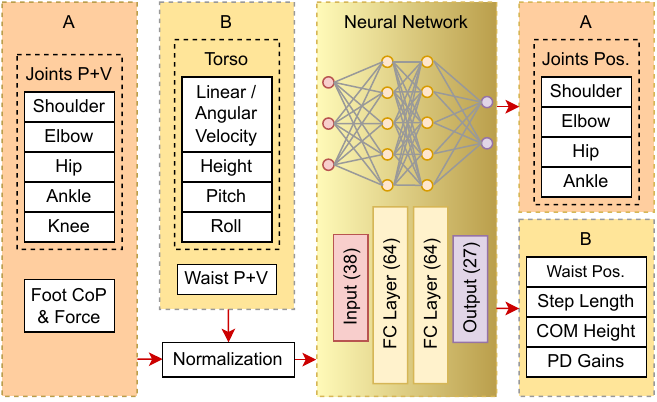}
    \caption{Network architecture, system space parameters and symmetry transformation groups used for data augmentation: reflection symmetry transformation (A) and no transformation or inversion (B).}
    \label{fig:netarch}
    \vspace{-5mm}
\end{figure}

\vspace{-3mm}
\subsection{Network Architecture}
The network architecture and system space parameters are depicted in Fig. \ref{fig:netarch}. The observations comprise the position of 6 joints: shoulder, hip and waist with 3 degrees of freedom (DoF), ankle with 2 DoF, knee and elbow with 1 DoF. All joints are mirrored except the waist. Additional observations include the foot relative centre of pressure (in $x$ and $y$) and respective force magnitude, the torso's linear and angular velocity, height, pitch, and roll; totalling 38 state variables. This data is fed to a neural network with 2 hidden layers of 64 neurons, that produces joint residuals, which are added to the precomputed trajectories; and high-level parameters to regulate the kernel pattern generator: step length, COM height, and two PD gain vectors ($\boldsymbol{K}_{\Phi}$ from \eqref{eq:PDTorso} and $\boldsymbol{K}_{\zeta}$ from \eqref{eq:dcm_tracker}). 

The system space parameters are grouped into two symmetry transformations categories for data augmentation. Category A includes duplicated observations that are mirrored, considering the sagittal plane. Category B includes unique observations that may remain unchanged (e.g. torso's height) or suffer an inversion transformation (e.g. roll angle).

\vspace{-3mm}
\subsection{Reward function}
The reward function tries to achieve one fundamental goal of balancing while keeping cyclic movement patterns. The balance goal seeks to keep the robot on its feet in all situations. The subgoal of performing cyclic movement patterns has the purpose of improving the human-like aspect of the behaviour. Specifically, it tries to reduce the neural network's influence (NNI) when there is no need to intervene. Both of these notions can be expressed through the following reward:
\vspace{-2mm}
\begin{equation}
R = 1 - \overbrace{\frac{1}{J}\sum_i^J \frac{|\delta_i|}{S_i}}^{NNI} ,\label{eq:reward} 
\end{equation}

\noindent where $\delta_i$ is the residual applied to joint position $i$, $J$ is the number of joints, and $S_i$ is the residual saturation value. It is important to note that the NNI component's goal is not to reduce energy consumption or range of motion, since it is only applied to the residuals and not the hybrid controller's output.

\section{Simulation Scenarios}  \label{sec:simScenarios}

To validate the performance of the proposed framework, a set of two learning scenarios and one test scenario has been designed. The goal of this structure is to prepare the physical robot to handle real world adverse conditions. We use the COMAN robot in PyBullet~\cite{coumans2020} -- an environment based on the open source Bullet Physics Engine. The simulated robot is 95 cm tall, weighs 31 kg, and has 23 joints (6 per leg, 4 per arm and 3 between the hip and the torso). 
\begin{figure*}[t]
 \centering

    \includegraphics[width = 0.96\linewidth,trim=0cm 0cm 0cm 0cm,clip]{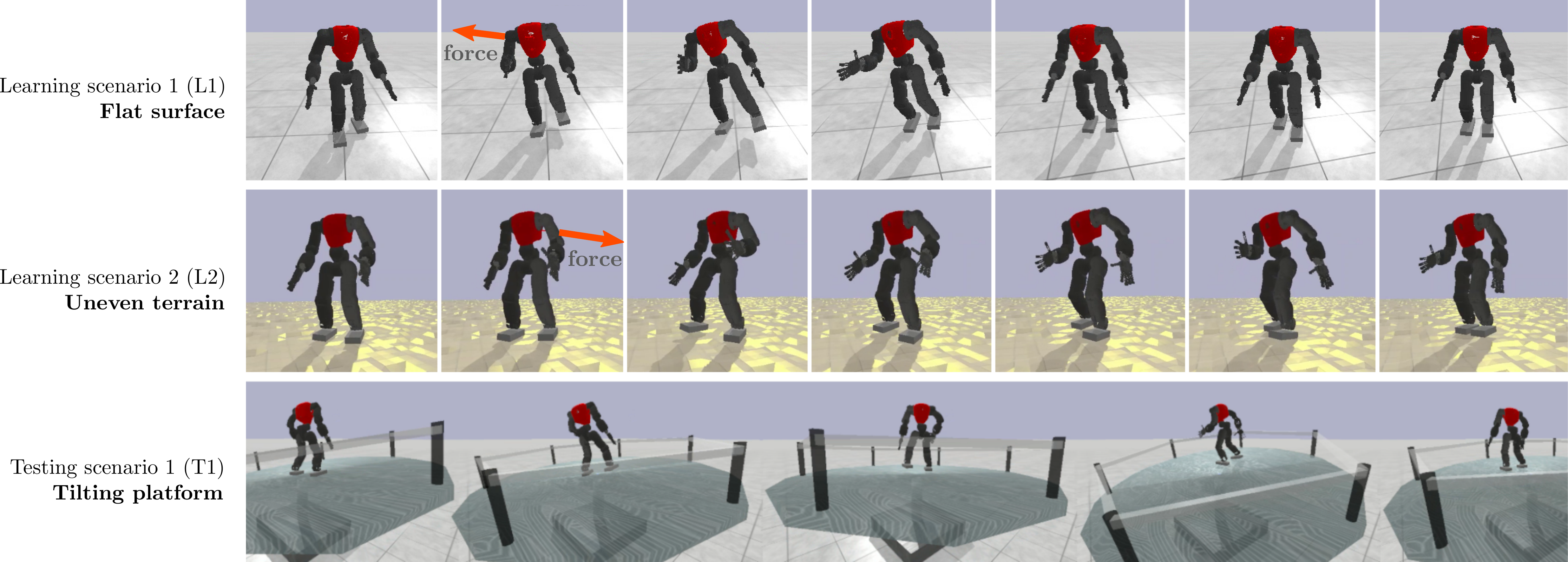}

    \caption{Simulation scenarios: The robot learns how to recover from external forces on a flat surface (\texttt{L1}) and in an uneven terrain with perturbations that can reach 2cm (\texttt{L2}); It was then tested on unseen scenarios on a tilting platform that moves erratically (\texttt{T1}).}
    \vspace{-4mm}
    \label{fig:simall}	
\end{figure*}

\subsection{Learning Scenario: flat terrain} \label{sec:scenarioL1}

The first learning scenario (\texttt{L1}) is composed of a flat platform (see Fig.~\ref{fig:simall}, top row), where the robot is initially placed in a neutral pose. It then starts to walk in place, while being pushed by an external force at random intervals, between 2.5 and 3.0 seconds. The force is applied for 25 ms and ranges from 500~N to 850~N. Its point of application is fixed at the torso's centre and its direction is determined randomly in the horizontal plane. The robot's objective is to avoid falling. The episode ends when the robot's height drops below 0.35~m.

\subsection{Learning Scenario: uneven terrain}

The second learning scenario (\texttt{L2}) is an extension of the first one, where the flat surface is replaced by an uneven terrain with perturbations that can reach 0.02~m, as depicted in Fig.~\ref{fig:simall}, middle row. The external force dynamics are the same.

\subsection{Testing Scenario: tilting platform}

The testing scenario (\texttt{T1}) was designed to evaluate the generalisation capabilities of the hybrid controller in unexpected circumstances. It is characterised by a tilting cylindrical platform (see Fig.~\ref{fig:simall}, bottom row), which is supported by two actuators that move on the $x$ and $y$ axes, and range between $-15\deg$ and $15\deg$. The position of each actuator is given by adding a random component $r\in[-8^\circ,8^\circ]$ to a correcting component $c=0.35\times P$, where $P$ is the position of the robot in the opposite axis to the actuator. The goal of the latter component is to keep the robot on top of the platform by encouraging it to move to the centre. The episode starts in a neutral state with the robot walking in place, and it ends when the robot falls, as in previous scenarios.

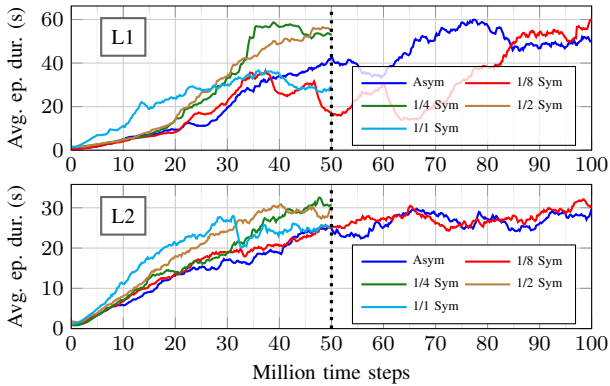
\begin{figure}[!t]

    \footnotesize
	\centering 
	\begin{tikzpicture}[]
	
	\pgfplotsset{
		height=3.5cm, width=8.5cm, compat=1.14,
	}

	\begin{axis}[
		ylabel=Avg. ep. dur. (s),
		legend pos=south east,
		legend cell align={left},
		legend style={font=\tiny, fill=white, fill opacity=0.8, draw opacity=1,text opacity=1},
		legend columns=2, 
		ytick distance= 20,
		xtick distance= 10,
        minor x tick num=1,
        ymin=0, xmin=0, xmax=100,
        grid=both,
        minor grid style={black!5},
        major grid style={black!20},
        minor x tick style={black!10},
	]
	\addplot[blue,mark=none,thick] table [x=x,y=s39,col sep=comma]{Plots/evo.csv};
	\addplot[red,mark=none, thick] table [x=x,y=s38,col sep=comma]{Plots/evo.csv};
	\addplot[green,mark=none,thick] table [x=x,y=s54,col sep=comma]{Plots/evo.csv};
	\addplot[brown,mark=none,thick] table [x=x,y=s51,col sep=comma]{Plots/evo.csv};
	\addplot[cyan,mark=none,thick] table [x=x,y=s47,col sep=comma]{Plots/evo.csv};
	\draw [dotted, very thick] (50,0) -- (50,65);
	\legend{Asym\\1/8 Sym\\1/4 Sym\\1/2 Sym\\1/1 Sym\\};
	\node[rectangle, draw=black!60, fill=white,  thick, minimum size=5mm, text width=4mm, align=center] at (10,52){L1};
	
	\end{axis}
	\end{tikzpicture}
	
	\begin{tikzpicture}[]
	
	\pgfplotsset{
		height=3.5cm, width=8.5cm, compat=1.14,
	}

	\begin{axis}[
		ylabel=Avg. ep. dur. (s),
		xlabel=Million time steps,
		legend pos=south east,
		legend cell align={left},
		legend style={font=\tiny},
		legend columns=2, 
		ytick distance= 10,
		xtick distance= 10,
        minor x tick num=1,
        ymin=0, xmin=0, xmax=100,
        grid=both,
        minor grid style={black!5},
        major grid style={black!20},
        minor x tick style={black!10},
	]
	\addplot[blue,mark=none,thick] table [x=x,y=s65,col sep=comma]{Plots/evo.csv};
	\addplot[red,mark=none, thick] table [x=x,y=s63,col sep=comma]{Plots/evo.csv};
	\addplot[green,mark=none,thick] table [x=x,y=s60,col sep=comma]{Plots/evo.csv};
	\addplot[brown,mark=none,thick] table [x=x,y=s57,col sep=comma]{Plots/evo.csv};
	\addplot[cyan,mark=none,thick] table [x=x,y=s59.64,col sep=comma]{Plots/evo.csv};
	\draw [dotted, very thick] (50,0) -- (50,35);
	\legend{Asym\\1/8 Sym\\1/4 Sym\\1/2 Sym\\1/1 Sym\\}
	\node[rectangle, draw=black!60, fill=white,  thick, minimum size=5mm, text width=4mm, align=center] at (10,28){L2};
		
	\end{axis}
	\end{tikzpicture}

	\caption{Learning curves for the models trained in scenario \texttt{L1} (top) and \texttt{L2} (bottom), under different symmetry configurations.}
	\label{fig:evo}
	\vspace{-3mm}
\end{figure}

\section{Simulations} \label{sec:simulations}
This section is focused on a set of assessments and analysis of the 
framework regarding overall performance, optimised policy behaviour, symmetry, and robustness.

\begin{figure*}[t]
 \centering
    \includegraphics[width = 0.96\linewidth, trim={2cm 0 0cm 0cm},clip]{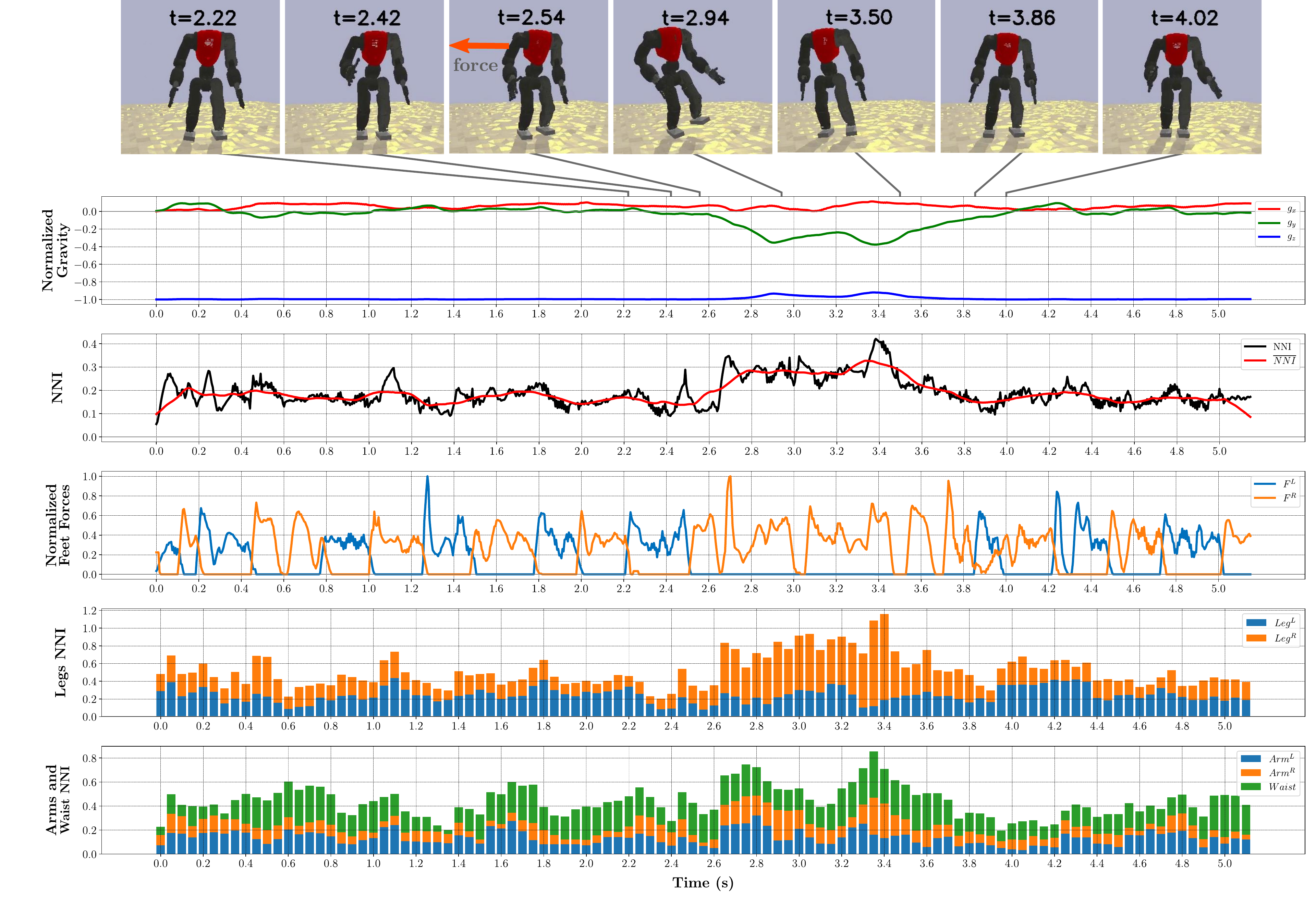}
\vspace{-3mm}
	\caption{ Analysis of simulation using the asymmetrical \texttt{L2} Model on the \texttt{L2} scenario for 5 seconds, with a single external push, applied to the robot's torso for 0.025s with a force of 850N. Seven snapshots before and after the push are presented. During entire simulation period, different metrics were sampled at 200Hz as: normalised gravity vector, relative to the robot's torso; global NNI, as defined in \eqref{eq:reward}; normalised feet forces; and NNI per joint group.}
	\label{fig:NNIAnalysis}       
	\vspace{-2mm}
\end{figure*} 
\vspace{-0mm}

\subsection{Performance analysis}
Five different symmetry ratios were tested per learning scenario, totalling ten different configurations. The symmetry ratios were 0 (no data augmentation), 1/8 (1 symmetrical sample is generated per 8 acquired samples), 1/4, 1/2 and 1/1 (full symmetry). For each configuration, five models were trained. Fig.~\ref{fig:evo} depicts the learning curves for the best model in each configuration. The results are grouped according to the training scenario (\texttt{L1} above and \texttt{L2} below). Most optimisations ran for 50M time steps. However, the asymmetric and 1/8 symmetry configurations needed 100M time steps to reach a plateau. For the configurations that included data augmentation, the best performing ratios were 1/4 and 1/2, with similar results. In a subjective visual evaluation, the 1/2 ratio model seems to be marginally better in producing a human-like behaviour. For the remainder of this section, we will compare in greater detail the asymmetric version with the 1/2 symmetric version. A video including the results is attached as supplementary material.

It is important to note that the average episode duration reported by these learning curves results from a stochastic policy with a non-negligible random component. To better assess the optimised models, they were tested in each scenario (including \texttt{T1} --- the only test scenario) for 1000 episodes using the corresponding deterministic policy. Moreover, to be fair with every approach, only the evolution until 50M time steps was considered in these tests. Table~\ref{tb:results} compares the average performance of 4 models against the baseline. The first four columns indicate, in this order, the episode duration, in seconds, in scenario \texttt{L1}, \texttt{L2} and \texttt{T1}; and the neural network influence (examined later in this section).

\begin{table}[!t]
    \centering
	\caption{Average results per learning configuration}
	\label{tb:results}
\begin{tabular}{l|c|c|c|c|c}
\multicolumn{1}{c|}{\multirow{2}{*}{\begin{tabular}[c]{@{}c@{}}Learning \\ configuration\end{tabular}}} & \multicolumn{3}{c|}{Episode duration (s)} & \multirow{2}{*}{\begin{tabular}[c]{@{}c@{}}N. Network\\ Influence\end{tabular}} & \multirow{2}{*}{\begin{tabular}[c]{@{}c@{}}M. Sym.\\ Index\end{tabular}} \\ \cline{2-4}
\multicolumn{1}{c|}{} & L1 & L2 & T1 &  &  \\ \hline
Baseline & 3.47 & 1.51 & 1.87 & - & - \\ \hline
L1 Asym & 104.5 & 5.1 & 4.8 & 0.072 & 1.42 \\ \hline
L1 1/2 Sym & 202.2 & 4.6 & 4.8 & 0.055 & 1.19 \\ \hline
L2 Asym & 321.9 & 34.2 & 27.8 & 0.165 & 1.23 \\ \hline
L2 1/2 Sym & 193.7 & 43.5 & 21.0 & 0.127 & 0.99
\end{tabular}
\vspace{-5mm}
\end{table}

The baseline version (without residuals) is not able to handle the strong external forces applied in scenario \texttt{L1}, falling on average after 3.47 s, which is typically after the first push. On \texttt{L2}, it falls almost immediately due to the floor perturbations, an outcome which is also seen in \texttt{T1}. All four learned models are a great improvement over the baseline. As expected, the last two models that learned on \texttt{L2} were able to generalise successfully when tested on \texttt{L1} or \texttt{T1}, and, on the opposite side, the models that learned on \texttt{L1} did not perform well in unforeseen circumstances. 

However, some results were not expected. During training, the symmetrically-enhanced models performed better but, while testing in distinct scenarios, the asymmetrical models generalised better. Another interesting result is that the asymmetrical \texttt{L1} model performed worse in its own scenario (104.5 s) than the asymmetrical \texttt{L2} model (321.9 s). 

An initial hypothesis to explain this outcome would be to assume that learning on an uneven terrain requires additional effort to maintain balance and, consequently, produces a better policy. In fact, considering that the robot is already pushed periodically, gravity acts as an additional external force when the robot is standing on a slope. On its own, this explanation is insufficient because the robot that learned on the flat surface could continue the optimisation process until it found the better policy. However, this would only be true if the reward was solely focused on raw performance.

To better understand this outcome, we need to analyse the NNI column of table \ref{tb:results}, whose metric is defined in~\eqref{eq:reward}. Since \texttt{L2} and \texttt{L2 Sym} require additional effort to counteract gravity when standing on a slope, the robot learned to sacrifice its immediate reward by applying larger residuals in order to avoid falling. Naturally, this is a trade-off between cyclic movement patterns and raw performance. Moreover, learning an asymmetrical behaviour can arguably be considered more complex, leading to a higher network influence, which may explain why it generalises better than the symmetrical models.

\vspace{-1mm}
\subsection{Optimised policy behaviour analysis}
To present more detail about the overall behaviour of the optimised models and to explain how they improve the robot's stability significantly, we selected the asymmetrical L2 model to represent all the optimised models and tested it on the L2 scenario for five seconds while recording all observations and actions~(200Hz). In this simulation, while the robot was walking in place, at t=2.54s, it was subjected to a 850N external push at its torso’s centre for 0.025s. The robot was able to counteract this force and regain its stability. A set of snapshots along with five important plots are depicted in Fig.~\ref{fig:NNIAnalysis}, including the normalised gravity vector and feet forces, and the NNI on different joint groups.

The first plot shows the normalised gravity vector, relative to the robot's torso. After applying the push, the robot leans considerably, with an inclination of 23 degrees, which can be characterised as a severe perturbation. Before the push, the average NNI~($\overline{\text{NNI}}$) is less than 0.2. The robot applies small corrections to keep its stability while walking in place on the uneven terrain. After triggering the external push, the network's influence rises 50\%, which translates into larger residuals, as a response to regain stability. After returning to a normal state, the NNI is smoothly reduced. These results validate the policy's objective stated in \eqref{eq:reward}, by adjusting the NNI according to the robot's requirements at a given moment.

To identify the distinct strategies at play, we broke down the network's influence into groups of limbs and waist, and chose feet forces as an additional metric. The total force acting perpendicular on each foot encodes the actual stride time and, by inspecting its plot, we can infer that even before the push, changing the stride time is one of the employed strategies.

The network's influence per group was obtained by applying the NNI formula from \eqref{eq:reward} to the joints in a given group. The 2 bottom plots represent the groups of joints associated with both legs, both arms, and waist, using a stacked bar chart, where each bar represents the mean of 10 control steps (0.05s).

During the push, the support leg had the most expressive response in comparison with the other joint groups. This behaviour is best understood by analysing a slow-motion video of the push (available as supplementary material).
The robot starts hopping on the support leg, while using its upper body as a complementary strategy to shift the COM by swinging the arms as required. This process persists until the robot is stable enough to return to the initial walking gait. These strategies, along with adjusting the stride time and COM height, allow the robot to perform seamless transitions, like humans would do unconsciously. 

\vspace{-3mm}
\subsection{Symmetry analysis}

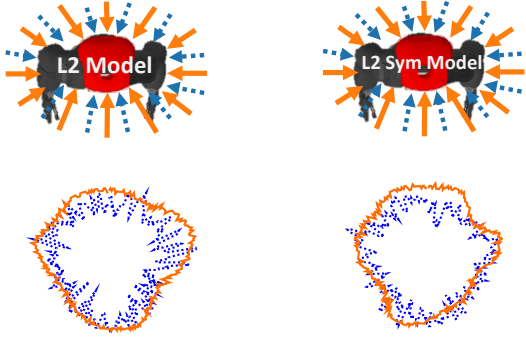
\begin{figure}[!t]
\begin{tabular}{c}
\hspace{6mm}
\includegraphics[width =0.7\linewidth, trim = 0cm 15cm 18.cm 0cm]{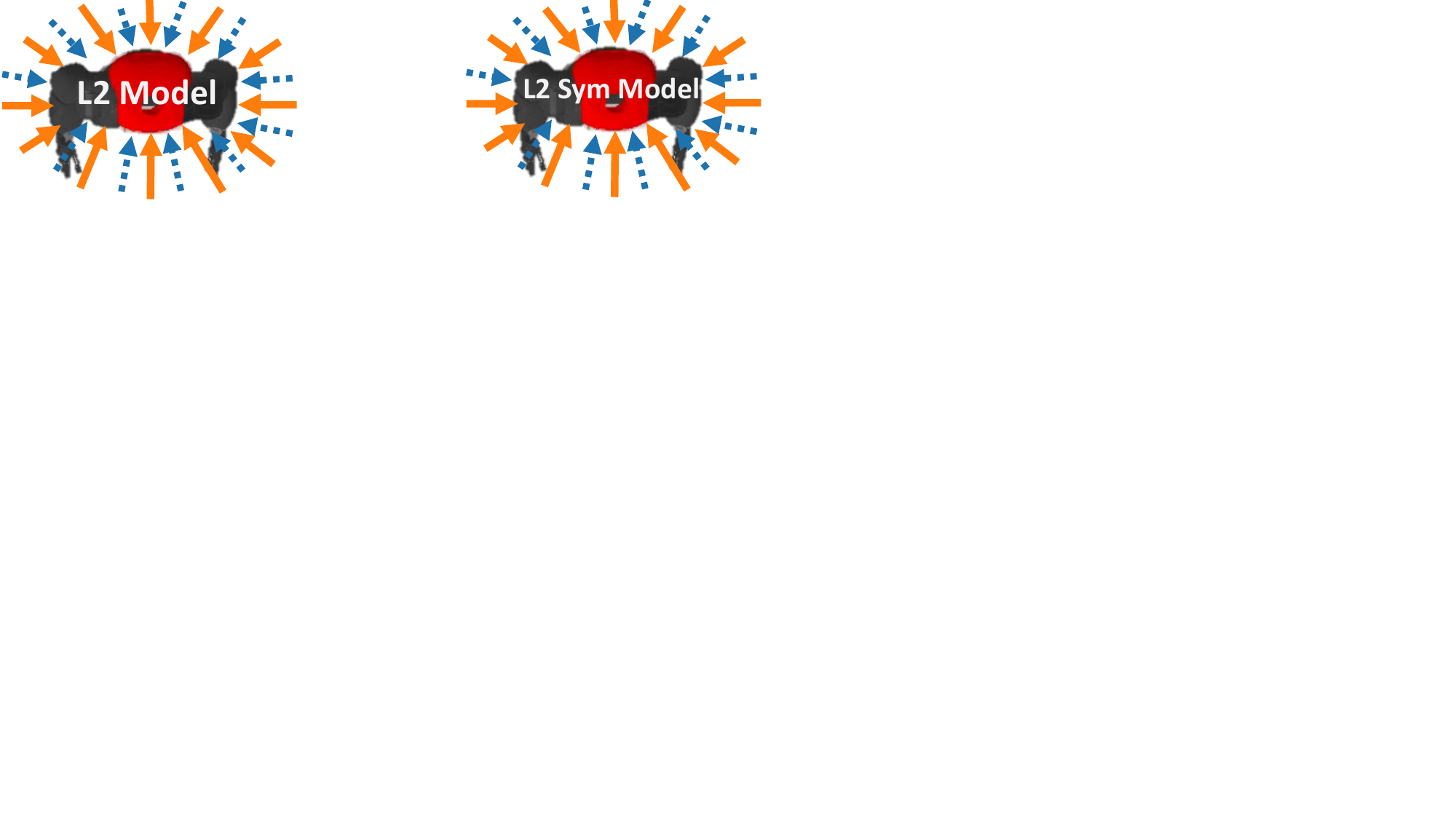}
      \\
 \hspace{-3.9cm}	
\begin{tikzpicture}
\tikzstyle{every node}=[font=\scriptsize]

\begin{polaraxis}[
  xticklabel=$\pgfmathprintnumber{\tick}^\circ$,
  yticklabel=$\pgfmathprintnumber{\tick}\operatorname{N}$,
  xtick={0,30,...,330},
  ytick={400,800,1200},
  ymin=400, ymax=1400,
  rotate=90,
  y coord trafo/.code=\pgfmathparse{#1-400},
  y coord inv trafo/.code=\pgfmathparse{#1+400},
  xticklabel style={anchor=\tick-90},
  yticklabel style={anchor=south, yshift=-0.15cm,fill=white, fill opacity=0.6, text opacity=1,font=\tiny},
  y axis line style={},
  ytick style={yshift=0cm},
  width=4.3cm,height=4.3cm,
  set layers = axis on top,
]

\addplot [no markers, thick, blue, densely dotted] table [col sep=comma, y=Force] {Plots/65_6.csv};
\addplot [no markers, thick, orange!85!red] table [col sep=comma, y=Force] {Plots/65_5.csv};

\end{polaraxis}


\end{tikzpicture}
\hspace{-0.2cm}
\begin{tikzpicture}
\tikzstyle{every node}=[font=\scriptsize]
\begin{polaraxis}[
  xticklabel=$\pgfmathprintnumber{\tick}^\circ$,
  yticklabel=$\pgfmathprintnumber{\tick}\operatorname{N}$,
  xtick={0,30,...,330},
  ytick={400,800,1200},
  ymin=400, ymax=1400,
  rotate=90,
  y coord trafo/.code=\pgfmathparse{#1-400},
  y coord inv trafo/.code=\pgfmathparse{#1+400},
  xticklabel style={anchor=\tick-90},
  yticklabel style={anchor=south, yshift=-0.15cm,fill=white, fill opacity=0.6, text opacity=1,font=\tiny},
  y axis line style={},
  ytick style={yshift=0cm},
  width=4.3cm,height=4.3cm,
  set layers = axis on top,
]
\addplot [no markers, thick, blue, densely dotted] table [col sep=comma, y=Force] {Plots/57_6.csv};
\addplot [no markers, thick, orange!85!red] table [col sep=comma, y=Force] {Plots/57_5.csv};
\end{polaraxis}
\end{tikzpicture}
\end{tabular}
\vspace{-1mm}
	\caption{ Maximum radially applied external force from which the robot can consistently recover as a function of the direction of application, where zero degrees corresponds to the front of the robot. On the left is shown the model which learned on \texttt{L2} and on the right \texttt{L2 Sym}. The force was applied both in the flat terrain (solid orange line) and the uneven terrain (dotted blue line). The radial y-axis range is $[400,1400]$N. The maximum withstood force was 1300N for the \texttt{L2} model in the flat terrain, at 290 degrees.}
	\vspace{-4mm}
	\label{fig:radial}
\end{figure}

Symmetry is an important property of human behaviours, often associated with positive reactions, as opposed to asymmetry \cite{evans2012human}. However, humans are not perfectly symmetrical, and unbalanced gait patterns can be perceived as unimpaired or normal, within reason \cite{handvzic15}. Therefore, in the context of human-like behaviours, the symmetry of a model should be leveraged, but not to the point where it becomes a hard constraint. In these simulations, the kernel pattern generator produces symmetrical trajectories upon which the neural network residuals are applied. To evaluate the residuals symmetry, we built upon the concept of Symmetry Index (SI) proposed by Robinson et. al~\cite{robinson1987}. The original method compares the kinematic properties of each lower limb. To address the issues caused by abstracting the kinematic properties of each joint, we propose the Mirrored Symmetry Index (MSI):

\begin{equation}
\mathrm{MSI} = \frac{\|\boldsymbol{\delta}_t-\boldsymbol{\delta}_t'\|_1}{0.5\times(\|\boldsymbol{\delta}_t\|_1+\|\boldsymbol{\delta}_t'\|_1)},
\label{eq:MSI}    
\end{equation}

\noindent where $\boldsymbol{\delta}_t=[\delta^t_1,...,\delta^t_n]$ is the vector of residuals applied to each joint during time step $t$, $\|\cdot\|_1$ is the $\ell$1-norm, and $\boldsymbol{\delta}_t'$ is the vector of residuals applied to the symmetric set of joints if the current state was also symmetrically transformed, i.e., $\boldsymbol{\delta}_t'\sim \pi(\cdot|f(S_t))$, where $\pi$ is a stochastic policy. Instead of evaluating an average kinematic feature, the MSI computes a symmetry index at each instant, which can then be averaged for a full trajectory to obtain a global symmetry assessment.

As seen in Table~\ref{tb:results}, the models which were learned using the data augmentation method obtained a lower MSI value, when compared to the other two models. The results do not show a large reduction, which can be explained by the analytical controller's role in regulating the trajectory symmetry, and the relaxed data augmentation restriction imposed to the network. 

To assess the notion of symmetry on a practical scenario, the models trained on \texttt{L2} and \texttt{L2 Sym} were subjected to a test where an external force with constantly increasing norm is radially applied to the robot in a given direction. When the robot is no longer able to recover consistently (more than 50\% of the trials), the maximum force is registered and another direction is tested. The result can be seen in Fig.~\ref{fig:radial} on the flat terrain (solid orange line) and uneven terrain (dotted blue line). In both cases, the robot is able to better withstand  forces that are applied to the front (around $0 \deg$). On one side, the symmetrically-enhanced version presents a more balanced result, which can be visually perceived. On the other side, the asymmetrical model can withstand larger forces around $300 \deg$. This difference consists of a trade-off between symmetry and raw performance.

\vspace{-2mm}
\subsection{Robustness}
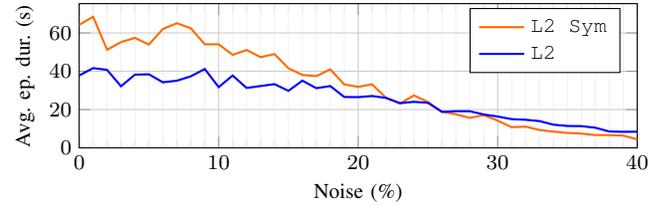
\begin{figure}[!t]
    \footnotesize
	\centering 
	\begin{tikzpicture}[]
	
	\pgfplotsset{
		height=3.5cm, width=9cm, compat=1.14,
	}

	\begin{axis}[
		ylabel=Avg. ep. dur. (s),
		xlabel=Noise (\%),
		legend pos=north east,
		legend cell align={left},
		ytick distance= 20,
		xtick distance= 10,
        minor x tick num=9,
        ymin=0, xmin=0, xmax=40,
        grid=both,
        minor grid style={black!5},
        major grid style={black!20},
        minor x tick style={black!10},
	]
	\addplot[orange!85!red,mark=none, thick] 
	table [x expr=\thisrowno{0}*100, y=duration, col sep=comma]{Plots/noise57uneven40.csv};
	\addplot[blue,mark=none,mark options={blue, scale=0.8}, thick] 
	table [x expr=\thisrowno{0}*100, y=duration, col sep=comma]{Plots/noise65uneven40.csv};
	
	\legend{\texttt{L2 Sym}\\\texttt{L2}\\}
	\end{axis}
	\end{tikzpicture}
	
    \vspace{-3mm}
	\caption{Average episode duration as a function of noise applied to the state observations for the symmetrical (orange line) and asymmetrical (blue line) models learned and tested on the uneven terrain.}
	\label{fig:noise}
	\vspace{-3mm}
\end{figure}

Finally, we present a robustness analysis, which is a matter of significant concern on real applications. To test this, the state variables are multiplied by a random factor that follows a uniform distribution $z\sim\mathcal{U}(1.0,N)$ where $N$ ranges from $1.0$ to $1.4$, i.e., $0\%$ to $40\%$ of maximum noise. Fig.~\ref{fig:noise} shows the average impact of this artificial perturbation on the average episode duration, on the uneven terrain scenario, while being pushed by an external force (described in Section~\ref{sec:scenarioL1}) with a fixed interval of 3.5 seconds. Both the symmetrical and asymmetrical models can withstand a maximum noise of $20\%$ without dropping below the 30 s mark, which attests the models' robustness in considerably noisy scenarios.

\begin{figure}[!t]
    \centering
    \includegraphics[width = 0.95\linewidth,trim = 0cm 9cm 11cm 0cm,clip]{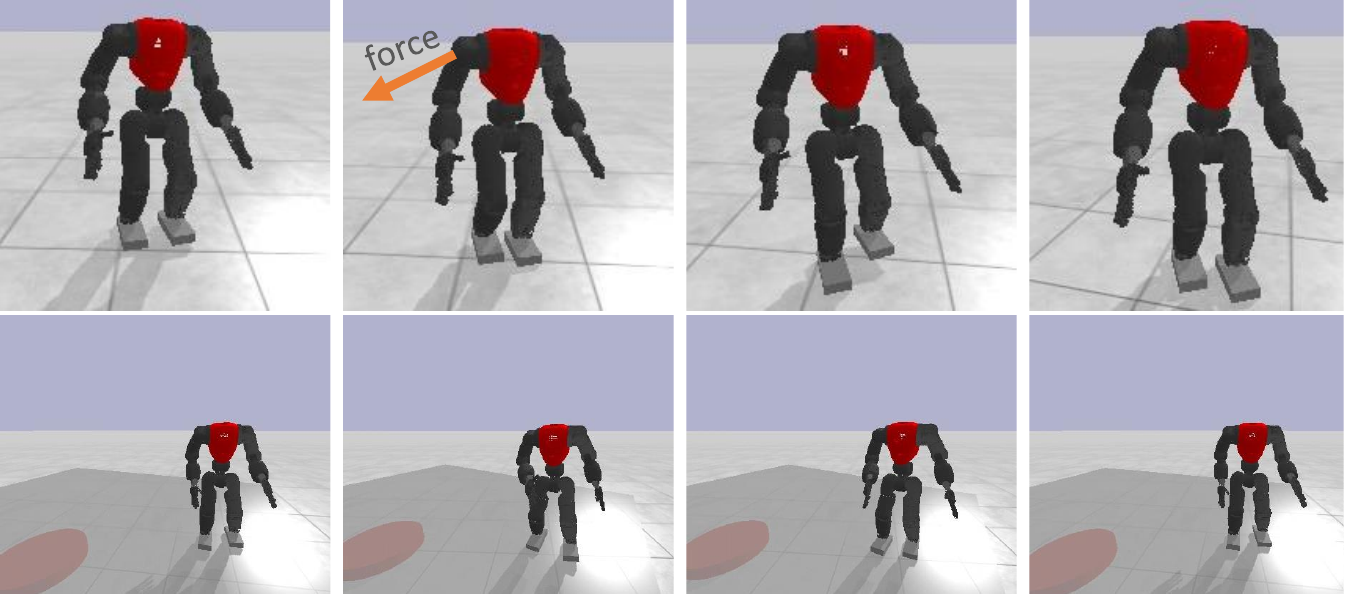}
    \vspace{-3mm}

    \caption{Models trained in this work are applied to  other gaits, such as walking forward and walking in place on a rotating platform.}
    \label{fig:walk}
\vspace{-4mm}
\end{figure}

\subsection{Applicability to Walking}
The models trained in this work were applied to different gaits, managing to attain a satisfactory performance while walking forward and being pushed, and while walking in place on a rotating platform (see Fig. \ref{fig:walk}). Changing direction or walking sideways can cause instability with the current configuration. However, these results, which are shown in the supplementary material, reveal a significant generalisation ability, considering that no model was trained specifically for this task.
\vspace{-1mm}

\section{Conclusion} \label{sec:conclusion}
In this paper, we proposed a locomotion framework based on a tight coupling between analytical control and deep reinforcement learning to combine the potential of both approaches. First, we developed a closed-loop fully parametric gait generation kernel. Then, we designed a learning framework which extends PPO with symmetric partial data augmentation to learn residuals dynamics. This hybrid approach aims at unlocking the full potential of the robot by exploiting the consistency of the analytical solution, the generalisation ability of neural networks, and the model's symmetry, while not totally constraining the exploration of asymmetric reactions. 

The results attest the models' robustness in considerably noisy environments. The symmetry enhanced models were able to perform better in the scenarios where they learned, and with less samples, but were not able to generalise as well in unforeseen circumstances. However, the difference is partially explained by the way the reward function's influence penalty is less restrictive in challenging conditions. Preliminary results show that the models trained in this work already generalise well to other gaits, such as walking forward and walking in place on a rotating platform. In the future, we would like to explore the application of this hybrid approach to other types of gait, including running and climbing. 

\vspace{-2mm}


\bibliographystyle{IEEEtran}
\bibliography{main}

\begin{thebibliography}{10}
\providecommand{\url}[1]{#1}
\csname url@rmstyle\endcsname
\providecommand{\newblock}{\relax}
\providecommand{\bibinfo}[2]{#2}
\providecommand\BIBentrySTDinterwordspacing{\spaceskip=0pt\relax}
\providecommand\BIBentryALTinterwordstretchfactor{4}
\providecommand\BIBentryALTinterwordspacing{\spaceskip=\fontdimen2\font plus
\BIBentryALTinterwordstretchfactor\fontdimen3\font minus
  \fontdimen4\font\relax}
\providecommand\BIBforeignlanguage[2]{{%
\expandafter\ifx\csname l@#1\endcsname\relax
\typeout{** WARNING: IEEEtran.bst: No hyphenation pattern has been}%
\typeout{** loaded for the language `#1'. Using the pattern for}%
\typeout{** the default language instead.}%
\else
\language=\csname l@#1\endcsname
\fi
#2}}

\bibitem{mason2016balancing}
S.~Mason, N.~Rotella, S.~Schaal, and L.~Righetti, ``Balancing and walking using
  full dynamics lqr control with contact constraints,'' in \emph{2016 IEEE-RAS
  16th International Conference on Humanoid Robots (Humanoids)}.\hskip 1em plus
  0.5em minus 0.4em\relax IEEE, 2016, pp. 63--68.

\bibitem{faraji2019bipedal}
S.~Faraji, H.~Razavi, and A.~J. Ijspeert, ``Bipedal walking and push recovery
  with a stepping strategy based on time-projection control,'' \emph{The
  International Journal of Robotics Research}, vol.~38, no.~5, pp. 587--611,
  2019.

\bibitem{caron2020icra}
\BIBentryALTinterwordspacing
S.~Caron, ``Biped stabilization by linear feedback of the variable-height
  inverted pendulum model,'' in \emph{IEEE International Conference on Robotics
  and Automation}, May 2020. [Online]. Available:
  \url{https://hal.archives-ouvertes.fr/hal-02289919}
\BIBentrySTDinterwordspacing

\bibitem{takenaka2009real}
T.~Takenaka, T.~Matsumoto, and T.~Yoshiike, ``Real time motion generation and
  control for biped robot-1st report: Walking gait pattern generation,'' in
  \emph{Intelligent Robots and Systems, 2009. IROS 2009. IEEE/RSJ International
  Conference on}.\hskip 1em plus 0.5em minus 0.4em\relax IEEE, 2009, pp.
  1084--1091.

\bibitem{englsberger2015three}
J.~Englsberger, C.~Ott, and A.~Albu-Sch{\"a}ffer, ``Three-dimensional bipedal
  walking control based on divergent component of motion,'' \emph{IEEE
  Transactions on Robotics}, vol.~31, no.~2, pp. 355--368, 2015.

\bibitem{morisawa2014biped}
M.~Morisawa, N.~Kita, S.~Nakaoka, K.~Kaneko, S.~Kajita, and F.~Kanehiro,
  ``Biped locomotion control for uneven terrain with narrow support region,''
  in \emph{System Integration (SII), 2014 IEEE/SICE International Symposium
  on}.\hskip 1em plus 0.5em minus 0.4em\relax IEEE, 2014, pp. 34--39.

\bibitem{8967778}
M.~{Kasaei}, N.~{Lau}, and A.~{Pereira}, ``A robust biped locomotion based on
  linear-quadratic-gaussian controller and divergent component of motion,'' in
  \emph{2019 IEEE/RSJ International Conference on Intelligent Robots and
  Systems (IROS)}, 2019, pp. 1429--1434.

\bibitem{brasseur2015robust}
C.~Brasseur, A.~Sherikov, C.~Collette, D.~Dimitrov, and P.-B. Wieber, ``A
  robust linear mpc approach to online generation of 3d biped walking motion,''
  in \emph{2015 IEEE-RAS 15th International Conference on Humanoid Robots
  (Humanoids)}.\hskip 1em plus 0.5em minus 0.4em\relax IEEE, 2015, pp.
  595--601.

\bibitem{kajita2018biped}
S.~Kajita, M.~Benallegue, R.~Cisneros, T.~Sakaguchi, S.~Nakaoka, M.~Morisawa,
  H.~Kaminaga, I.~Kumagai, K.~Kaneko, and F.~Kanehiro, ``Biped gait control
  based on spatially quantized dynamics,'' in \emph{2018 IEEE-RAS 18th
  International Conference on Humanoid Robots (Humanoids)}.\hskip 1em plus
  0.5em minus 0.4em\relax IEEE, 2018, pp. 75--81.

\bibitem{seyde2018inclusion}
T.~Seyde, A.~Shrivastava, J.~Englsberger, S.~Bertrand, J.~Pratt, and R.~J.
  Griffin, ``Inclusion of angular momentum during planning for capture point
  based walking,'' in \emph{2018 IEEE International Conference on Robotics and
  Automation (ICRA)}.\hskip 1em plus 0.5em minus 0.4em\relax IEEE, 2018, pp.
  1791--1798.

\bibitem{gu2017deep}
S.~Gu, E.~Holly, T.~Lillicrap, and S.~Levine, ``Deep reinforcement learning for
  robotic manipulation with asynchronous off-policy updates,'' in \emph{2017
  IEEE international conference on robotics and automation (ICRA)}.\hskip 1em
  plus 0.5em minus 0.4em\relax IEEE, 2017, pp. 3389--3396.

\bibitem{abreu2019learning}
M.~Abreu, N.~Lau, A.~Sousa, and L.~P. Reis, ``Learning low level skills from
  scratch for humanoid robot soccer using deep reinforcement learning,'' in
  \emph{2019 IEEE International Conference on Autonomous Robot Systems and
  Competitions (ICARSC)}.\hskip 1em plus 0.5em minus 0.4em\relax IEEE, 2019,
  pp. 1--8.

\bibitem{lin2020robotics}
Y.~{Lin}, J.~{Huang}, M.~{Zimmer}, Y.~{Guan}, J.~{Rojas}, and P.~{Weng},
  ``Invariant transform experience replay: Data augmentation for deep
  reinforcement learning,'' \emph{IEEE Robotics and Automation Letters},
  vol.~5, no.~4, pp. 6615--6622, 2020.

\bibitem{abdolhosseini2019learning}
F.~Abdolhosseini, H.~Y. Ling, Z.~Xie, X.~B. Peng, and M.~van~de Panne, ``On
  learning symmetric locomotion,'' in \emph{Motion, Interaction and Games},
  2019, pp. 1--10.

\bibitem{mishra2019augmenting}
S.~Mishra, A.~Abdolmaleki, A.~Guez, P.~Trochim, and D.~Precup, ``Augmenting
  learning using symmetry in a biologically-inspired domain,'' arXiv preprint
  arXiv:1910.00528, 2019.

\bibitem{Abreu2019runfaster}
M.~Abreu, L.~P. Reis, and N.~Lau, ``{Learning to Run Faster in a Humanoid Robot
  Soccer Environment Through Reinforcement Learning},'' in \emph{RoboCup 2019:
  Robot World Cup XXIII}, S.~Chalup, T.~Niemueller, J.~Suthakorn, and M.-A.
  Williams, Eds.\hskip 1em plus 0.5em minus 0.4em\relax Cham: Springer
  International Publishing, 2019, pp. 3--15.

\bibitem{koryakovskiy2018model}
I.~Koryakovskiy, M.~Kudruss, H.~Vallery, R.~Babu{\v{s}}ka, and W.~Caarls,
  ``Model-plant mismatch compensation using reinforcement learning,''
  \emph{IEEE Robotics and Automation Letters}, vol.~3, no.~3, pp. 2471--2477,
  2018.

\bibitem{ahn2020data}
J.~Ahn, J.~Lee, and L.~Sentis, ``Data-efficient and safe learning for humanoid
  locomotion aided by a dynamic balancing model,'' \emph{IEEE Robotics and
  Automation Letters}, vol.~5, no.~3, pp. 4376--4383, 2020.

\bibitem{yang2018learning}
C.~Yang, K.~Yuan, W.~Merkt, T.~Komura, S.~Vijayakumar, and Z.~Li, ``Learning
  whole-body motor skills for humanoids,'' in \emph{2018 IEEE-RAS 18th
  International Conference on Humanoid Robots (Humanoids)}.\hskip 1em plus
  0.5em minus 0.4em\relax IEEE, 2018, pp. 270--276.

\bibitem{schulman2017ppo}
J.~Schulman, F.~Wolski, P.~Dhariwal, A.~Radford, and O.~Klimov, ``Proximal
  policy optimization algorithms,'' arXiv preprint arXiv:1707.06347, 2017.

\bibitem{ravindran2001}
B.~Ravindran and A.~G. Barto, ``Symmetries and model minimization in markov
  decision processes,'' USA, 2001.

\bibitem{sutton2018reinforcement}
R.~S. Sutton and A.~G. Barto, \emph{Reinforcement learning: An
  introduction}.\hskip 1em plus 0.5em minus 0.4em\relax MIT press, 2018.

\bibitem{schulman2018gae}
J.~Schulman, P.~Moritz, S.~Levine, M.~Jordan, and P.~Abbeel, ``High-dimensional
  continuous control using generalized advantage estimation,'' \emph{CoRR},
  vol. 1506.02438, 2018.

\bibitem{coumans2020}
E.~Coumans and Y.~Bai, ``Pybullet, a python module for physics simulation for
  games, robotics and machine learning,'' \url{http://pybullet.org},
  2016--2020.

\bibitem{evans2012human}
D.~W. Evans, P.~T. Orr, S.~M. Lazar, D.~Breton, J.~Gerard, D.~H. Ledbetter,
  K.~Janosco, J.~Dotts, and H.~Batchelder, ``Human preferences for symmetry:
  Subjective experience, cognitive conflict and cortical brain activity,''
  \emph{PLoS ONE}, vol.~7, no.~6, 2012.

\bibitem{handvzic15}
I.~Hand{\v{z}}i{\'c} and K.~B. Reed, ``Perception of gait patterns that deviate
  from normal and symmetric biped locomotion,'' \emph{Frontiers in psychology},
  vol.~6, p. 199, 2015.

\bibitem{robinson1987}
R.~Robinson, W.~Herzog, and B.~M. Nigg, ``Use of force platform variables to
  quantify the effects of chiropractic manipulation on gait symmetry,''
  \emph{Journal of manipulative and physiological therapeutics}, vol.~10,
  no.~4, pp. 172--176, 1987.

\end{thebibliography}

\end{document}